\documentclass[11pt]{article}
\usepackage{acl2016}
\usepackage{times}
\usepackage{latexsym}

\usepackage{url}

\usepackage{times}
\usepackage{helvet}
\usepackage{courier}

\usepackage{graphicx}
\usepackage{latexsym}
\usepackage{url}
\usepackage{amsmath}
\usepackage{amsfonts}
\usepackage{multirow}

\aclfinalcopy 

\title{Leveraging Lexical Resources for Learning Entity Embeddings \\
in Multi-Relational Data}

\author{Teng Long, Ryan Lowe, Jackie Chi Kit Cheung \& Doina Precup\\
School of Computer Science\\
McGill University\\
\texttt{teng.long@mail.mcgill.ca}\\
\texttt{\{ryan.lowe,jcheung,dprecup\}@cs.mcgill.ca} }

\begin{document}

\maketitle

\begin{abstract}

Recent work in learning vector-space embeddings for multi-relational data 
has focused on combining relational information derived from knowledge bases
with distributional information derived from large text corpora. 
We propose a simple approach that leverages the descriptions of entities or phrases available
in lexical resources, in conjunction with distributional semantics, in order
to derive a better initialization for training relational models. Applying this initialization to the TransE model results in significant new state-of-the-art performances on the WordNet dataset, decreasing the mean rank from the previous best of 212 to 51. It also results in faster convergence of the entity representations.  We find that there is a trade-off between improving the mean rank and the hits@10 with this approach. This illustrates that much remains to be understood regarding performance improvements in relational models.

\end{abstract}

\section{Introduction}

A surprising result of work on vector-space word embeddings is that word representations that are learned from a large training corpus display semantic regularities in the form of linear vector translations. For example, \newcite{mikolov2013linguistic} show that using their induced word vector representations, $\textit{king} - \textit{man} + \textit{woman} \approx \textit{queen}$. Such a structure is appealing because it provides an interpretation to the distributional vector space through lexical-semantic analogical inferences.

\begin{table}[t]
\begin{center}
\begin{tabular}{l c c c}

  &
&\multicolumn{1}{c}{\bf W2V } &\multicolumn{1}{c}{\bf GloVe }
\\ 
\multicolumn{1}{c}{\bf Dataset}& \multicolumn{1}{c}{\bf Total} &\textbf{found\% }& \textbf{found\% }\\ \hline
WN & 40943 & 9.7\% & 51.3\% \\
FB15k & 14951 & 4.0\% & 20.3\%
\end{tabular}
\end{center}
\caption{\label{missingembeddings} The percentage of WN and FB15k entities that can be found in the pre-trained word2vec (W2V) and GloVe vectors. This does not include the W2V
embeddings trained with the FB15k vocabulary\footnotemark, which covers 
93\% of the FB15k entities.}
\end{table}

Concurrent to that work, \newcite{bordes2013translating} proposed \textit{translating embeddings} (TransE), which takes a pre-existing semantic hierarchy as input and embeds its structure into a vector space. In their model, the linear relationship between two entities that are in some semantic relation to each other is an explicit part of the model's objective function. For example, given a relation such as $\textit{won}(\textit{Germany}, \textit{FIFA Worldcup})$, the TransE model learns vector representations for \textit{won}, \textit{Germany}, and \textit{FIFA Worldcup} such that $\textit{Germany} + \textit{won} \approx \textit{FIFA Worldcup}$.

\footnotetext{This means that word2vec was trained in the usual way on a large textual corpus, but the vocabulary was truncated to include as many entities from Freebase as possible. Indeed, this is the reason for the small overlap between W2V, GloVe, and the relational databases: after training the word embeddings, the vocabulary must be truncated to a reasonable size, which leaves out many entities from these datasets.}

A natural next step is to attempt to integrate the two approaches in order to develop a representation that is informed by both unstructured text and a structured knowledge base \cite{faruqui2014retrofitting,xu-etal-2014,fried-duh-2014,yang2014embedding}. However, existing work makes a crucial assumption---that reliable distributional vectors are available for all of the entities in the hierarchy being modeled. Unfortunately, this assumption does not hold in practice; when moving to a new domain with a new knowledge base, for example, there will likely be many entities or phrases for which there is no distributional information in the training corpus. This important problem is illustrated in Table~\ref{missingembeddings}, where most of the entities from WordNet and Freebase are seen to be missing from the distributional vectors derived using Word2Vec and GloVe trained on the Google News corpus. Even when the entities are found, they may not have occurred enough times in the training corpus for their vector representation to be reliable. What is needed is a method to derive entity representations that works well for both common and rare entities.

Fortunately, knowledge bases typically contain a short description or definition for each of the entities or phrases they contain. 
For example, in a medical dataset with many technical words, the Wikipedia pages, dictionary definitions, or medical descriptions via a site such as \texttt{medilexicon.com} could be leveraged as lexical resources. Similarly, when building language models for social media, resources such as \texttt{urbandicionary.com} could be used for information about slang words. 
For the WordNet and Freebase datasets, we use \textit{entity descriptions} which are readily available (see Table \ref{entdesc}). 

In this paper, we propose a simple and efficient procedure to convert these short descriptions into a vector space representation, with the help of existing word embedding models. These vectors are then used as the input to further training with the TransE model, in order to incorporate structural information. Our method provides a better initialization for the TransE model, not just for the entities that do not appear in the data, but in fact for \emph{all} entities. This is demonstrated by achieving state-of-the-art mean rank on an entity ranking task on two very different data sets: WordNet synsets with lexical semantic relations \cite{miller1995wordnet}, and Freebase named entities with general semantic relations \cite{bollacker2008freebase}.

\begin{table}[t]
\footnotesize
\begin{center}
\begin{tabular}{|l|l|}
  \hline
  \multicolumn{2}{|l|}{\textbf{WordNet Descriptions}} \\ \hline
  photography\#3 &\parbox[t]{5cm}{\raggedright \textit{the occupation of taking and printing photographs or making movies}} \\ \hline
  transmutation\#2 & \textit{a qualitative change} \\ \hline \hline
  \multicolumn{2}{|l|}{\textbf{Freebase Descriptions}} \\ \hline
  Stephen Harper & \parbox[t]{5cm}{\raggedright \textit{Stephen Joseph Harper is a Canadian politician who is the 22nd and current Prime Minister of Canada and the Leader of the Conservative Party...} }\\ \hline
  El Paso & \parbox[t]{5cm}{\raggedright \textit{El Paso is the county seat of El Paso County, Texas, United States, and lies in far West Texas...}} \\ \hline
\end{tabular}
\end{center}
\caption{\label{entdesc} Sample entity descriptions from WordNet and Freebase. As Freebase descriptions are lengthy paragraphs, only the first sentence is shown.}
\end{table}

\section{Related Work}

Dictionary definitions were the core component of early methods in word sense disambiguation (WSD), such as the Lesk algorithm~\nocite{lesk-1986} (1986). Chen et al. \shortcite{chen2014unified} build on the use of synset glosses for WSD by leveraging lexical resources. Our work goes further to tie these glosses together with relational semantics, a connection that has not been drawn in the literature before. The integration of lexical resources into distributional semantics has been studied in other lexical semantic tasks, such as synonym expansion~\cite{sinha2009combining}, relation extraction~\cite{kambhatla2004combining}, and calculating the semantic distance between concepts~\cite{mohammad2008measuring,marton2009estimating}. We aim to combine lexical resources and other semantic knowledge, but we do so in the context of neural network-based word embeddings, rather than in specific lexical semantic tasks.


\newcite{bordes2011learning} propose the Structured Embeddings (SE) model, which embeds entities into vectors and relations into matrices. The relation connection between two entities is modeled by the projection of their embeddings  into a different vector space.
\newcite{rothe2015autoextend} use Wordnet as a lexical resource to learn embeddings for synsets and lexemes.  
Perhaps most related to our work are previous relational models that
initialize their embeddings via distributional semantics calculated from a 
larger corpus. \newcite{socher2013reasoning} propose the Neural 
Tensor Network (NTN), and \newcite{yang2014embedding} the Bilinear
model using this technique.
Other approaches modify the objective function or change the structure of the
model in order to integrate distributional and relational information 
\cite{xu-etal-2014,fried-duh-2014,toutanova2015observed}. \newcite{faruqui2014retrofitting} retrofit word vectors \emph{after} they are trained according to
distributional criteria. We propose a method that does not necessitate post-processing of the embeddings, and can be applied orthogonally to the previously mentioned improvements.




\section{Architecture of the Approach}

\subsection{The TransE Model}

The Translating Embedding (TransE) model~\cite{bordes2013translating} has become one of the most popular multi-relational models due to its relative simplicity, scalability to large datasets, and (until recently) state-of-the-art results. It assumes a simple additive interaction between vector representations of entities and relations. More precisely, assume a given relationship triplet $(h,l,t)$ is valid; then, the embedding of the object $t$ should be very close to the embedding of the subject $h$ plus some vector in $\mathbb{R}^{k}$ that depends on the relation $l$\footnote{Note that we use $h, l, t \in \mathbb{R}^k$ to denote both the entities and relations, in addition to the vector representations of the entities and relations}.

For each positive triplet $(h, l, t) \in S$, a negative triplet $(h', l, t') \in S'$ is constructed by randomly sampling an entity from $E$ to replace either the subject $h$ or the object $t$ of the relationship. The training objective of TransE is to minimize the dissimilarity measure $d(h + l, t)$ of a positive triplet while ensuring that $d(h' + l, t')$ for the corrupted triplet remains large. 
This is accomplished by minimizing the hinge loss over the training set:
\begin{equation*}
L = \sum_{(h, l, t) \in S} \sum_{(h', l, t') \in S'}
	[\gamma + d(h + l, t) - d(h' + l, t')]_{+}
\end{equation*}
where $\gamma$ is the hinge loss margin and $[x]_{+}$ represents the positive portion of $x$. 
There is an additional constraint that the $L_{2}$-norm of entity embeddings 
(but not relation embeddings) must be 1, which prevents the training process to 
trivially minimize $L$ by artificially increasing the norms of entity
embeddings.

\subsection{Initializing Representations with Entity Descriptions}

We propose to leverage some external lexical resource to improve the quality of the entity vector representations. In general, this could consist of product descriptions in a product database, or information from a web resource.
For the WordNet and Freebase datasets, we use \textit{entity descriptions} which are readily available. 

Although there are many ways to incorporate this, we propose a simple method whereby the entity descriptions are used to \textit{initialize} the entity representations of the model, which we show to have empirical benefits. In particular, we first decompose the description of a given entity into a sequence of word vectors, and combine them into a single embedding by averaging. We then reduce the dimensionality using principle component analysis (PCA), which we found  experimentally to reduce overfitting. We obtain these word vectors using distributed representations computed using word2vec~\cite{mikolov2013efficient} and GloVe~\cite{pennington2014glove}. Approximating compositionality by averaging vector representations is simple, yet has some theoretical justification~\cite{tian2015mechanism} and can work well in practice~\cite{wieting2015towards}. 

Additional decisions need to be made concerning which parts of the entity description to include. In particular, if an entity description or word definition is longer than several sentences, using the entire description could cause a `dilution' of the desired embedding, as not all sentences will be equally pertinent. We solve this by only considering the first sentence of any entity description, which is often the most relevant one. This is necessary for Freebase, where the description length can be several paragraphs.

\begin{table*}[t]
\footnotesize
\begin{center}
\begin{tabular}{ |c l||c|c c|c c||c|c c|c c| }
   \hline 
  \multicolumn{2}{|c||}{\multirow{3}{*}{ }} 
  & \multicolumn{5}{|c||}{\textbf{WN}} &  \multicolumn{5}{c|}{\textbf{FB15k}}\\
  \cline{3-12}
  &&\multirow{2}{0pt}{$k$}& \multicolumn{2}{c|}{\textbf{Mean rank}}
  & \multicolumn{2}{c||}{\textbf{Hits@10}}& \multirow{2}{0pt}{$k$} &\multicolumn{2}{c|}{\textbf{Mean rank}}
  & \multicolumn{2}{c|}{\textbf{Hits@10}}\\
  \cline{4-7}\cline{9-12}
  &&& \textbf{Raw} & \textbf{Filt} & \textbf{Raw} & \textbf{Filt}& &\textbf{Raw} & \textbf{Filt} & \textbf{Raw} & \textbf{Filt} \\
  \hline 
\parbox[t]{2mm}{\multirow{5}{*}{\rotatebox[origin=c]{90}{\scriptsize{Prev. models}}}} & SE \cite{bordes2011learning}& ---& 1,011 & 985 & 68.5\% & 80.5\%& ---& 273 & 162  & 28.8\%  & 39.8\% \\ 
 
  &TransD (unif)  \cite{jiknowledge2015} & ---& 242 & 229 & 79.2\% & \textbf{92.5\%}& ---& 211 & \textbf{67} & 49.4\%  & 74.2\% \\ 
  &TransD (bern)  \cite{jiknowledge2015}  & ---& 224 & 212 & \textbf{79.6\%} & 92.2\%& ---& 194 & 91 & \textbf{53.4\%} &\textbf{ 77.3\%}\\ \cline{2-12} 
  
  &TransE random init.  & 20& 266 & 254& 76.1\%  & 89.2\%& 50& 195 & 92& 41.2\%   & 55.2\% \\ 
  &TransE Freebase W2V init.  &---&---&---&---&---& 50& 195& 91 & 41.3\%   & 55.4\% \\ \hline 
  
 \parbox[t]{2mm}{\multirow{4}{*}{\rotatebox[origin=c]{90}{\scriptsize{Our models}}}} &TransE W2V entity defs. (NS) & 30& 210 & 192& 78.5\%   & 92.1\%& 55& 195& 91 & 41.6\%  & 55.7\% \\ 
  &TransE GloVe entity defs. (NS)  & 30& \textbf{63} & \textbf{51}& 64.6\%  & 73.2\%& 55& 194 &  90 & 41.7\%  & 55.8\% \\   
  &TransE W2V entity defs. & 30& 191 & 179 & 77.8\%   & 91.6\%& 55& 195& 91 & 41.6\%  & 55.6\%  \\ 
  &TransE GloVe entity defs. &  30& 71 & 59& 75.3\%  & 88.0\%& 55&  \textbf{193} & 90 & 41.8\%  & 55.8\% \\ \hline
\end{tabular}
\end{center}
\caption{\label{results}  Comparison between random initialization and using the entity descriptions. `NS' tag indicates stopword removal from the entity descriptions
`TransE Freebase W2V init' model uses word2vec pre-trained with the Freebase vocabulary, and thus was not tested on WN.
}
\end{table*}

\begin{figure*}[]
\centering
\includegraphics[width=.43\textwidth]{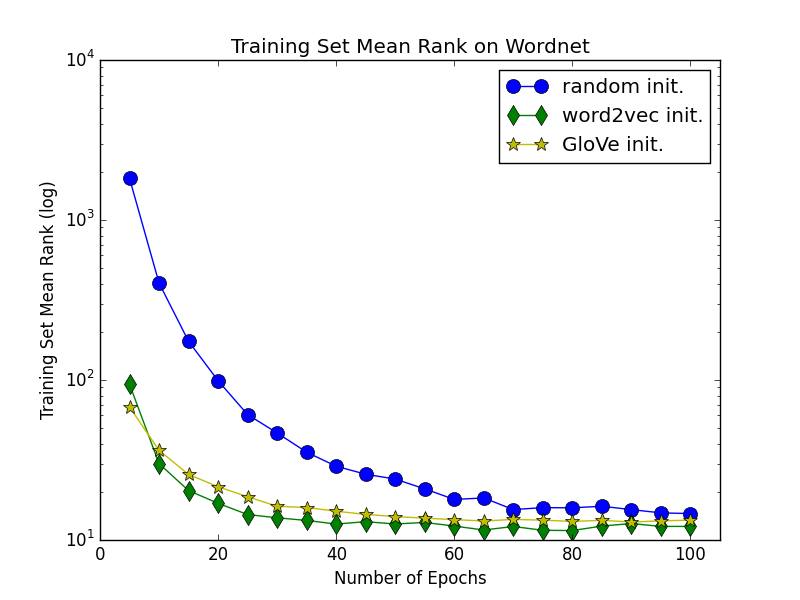}
\includegraphics[width=.43\textwidth]{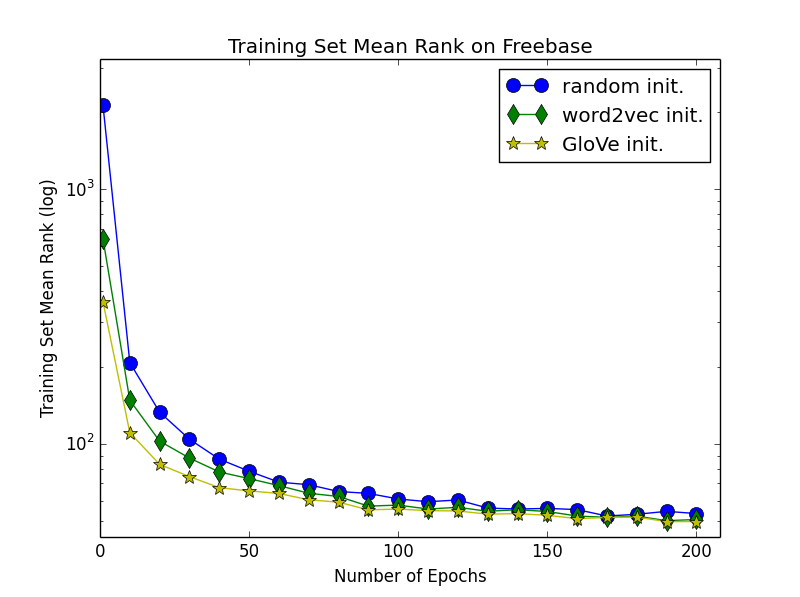}
\caption{Learning curves for the mean ranks on the training set for WordNet (left) and Freebase (right). }
\label{fig1}
\end{figure*}

\section{Experiments}

\subsection{Training and Testing Setup}

We perform experiments on the WordNet (WN)~\cite{miller1995wordnet} and Freebase (FB15k)~\cite{bollacker2008freebase} datasets 
used by the original
TransE model.  TransE hyperparameters include the learning rate $\lambda$ for stochastic gradient descent, the margin $\gamma$ for the hinge loss, the dimension of the embeddings $k$, and the dissimilarity metric $d$. 
For the TransE model with random initialization, we use the optimal hyperparameters from~\cite{bordes2013translating}: for WN, $\lambda = 0.01$, $\gamma = 2$, $k = 20$, and $d = L_{1}$-norm; for FB15k, $\lambda = 0.01$, $\gamma = 0.5$, $k = 50$, and $d = L_{2}$-norm. 
The values of $k$ were further tested to ensure that $k=20$ and $k=50$ were optimal. 
For the TransE model with strategic initialization, we used different embedding dimensions. The distributional vectors used in the entity descriptions are of dimension 1000 for the word2vec vectors with Freebase vocabulary, and dimension 300 in all other cases. 
Dimensionality reduction with PCA was then applied to reduce this to $k=30$ for WN, and $k=55$ for FB15k, which were empirically found to be optimal. PCA was necessary in this case as pre-trained vectors from word2vec and GloVe are not available for all dimension values. 

We use the same train/test/validation split and evaluation procedure as~\cite{bordes2013translating}: for each test triplet \textit{(h, l, t)}, we remove entity \textit{h} and \textit{t} in turn, and rank each entity in the dictionary by similarity according to the model.
We evaluate using the original and most common metrics for relational models: \textit{i)} the \textit{mean} of the predicted ranks, and \textit{ii)} \textit{hits@10}, which represents the percentage of correct entities found in the top 10 list; however, other metrics are possible, such as mean reciprocal rank (MRR). We evaluate in both the filtered setting, where other correct responses are removed from the lists ranked by the model, and the raw setting, where no changes are made.

We compare against the TransE model with random initialization, and the SE model \cite{bordes2011learning}. We also compare against the state-of-the-art TransD model~\cite{jiknowledge2015}. This model uses two vectors to represent each entity and relation; one to represent the meaning of the entity, and one to construct a mapping matrix dynamically. This allows for the representation of more diverse entities.

\subsection{Results and Analysis}

Table~\ref{results} summarizes the experimental results, compared to baseline and state-of-the-art relational models. We see that the mean rank is greatly improved for the TransE model with strategic initialization over random initialization. More surprisingly, all of our models achieve state-of-the-art performance for both raw and filtered data, compared to the recently developed TransD model. 
These results are highly significant with $p<10^{-3}$ according to the Mann-Whitney U test. Thus, even though our method is simple and straightforward to apply, it can still beat all attempts at more complicated structural modifications to the TransE model on this dataset. Further, the fact that our optimal embedding dimensions are larger (30 and 55 vs. 20 and 50) suggests that our initialization approach helps avoid overfitting.

For Freebase, our models slightly outperform the TransE model with random initialization, with p-values of 0.173 and 0.410 for initialization with descriptions (including stopwords) using GloVe and word2vec, respectively. We also see improvements over the case of direct initialization with word2vec. Further, we set a new state-of-the-art for mean rank on the raw data, though the improvement is marginal. 

Finally, we see in Figure \ref{fig1} that the TransE model converges more quickly during training when initialized with our approach, compared to random initialization. This is particularly true on WordNet.

\paragraph{Mean rank and hits@10 discrepancy} It is interesting to note the relationship between the mean rank and hits@10. By changing our model, we are able to increase one at the expense of the other. For example, using word2vec without stopwords gives similar hits@10 to TransD with better mean rank, while using GloVe further improves the mean rank at a cost to hits@10. The exact nature of this trade-off isn't clear, and is an interesting avenue for future work.

However, there are potential reasons for the results discrepancy betweeen mean rank and hits@10. 
We conjecture that our model helps avoid `disasters' where some correct entities are ranked very low. For TransE with random initialization, these disasters cause a large decrease in mean rank, which is significantly improved by our model. On the other hand, reducing the number of correct entities that are poorly ranked may not significantly affect the hits@10, since this only considers entities near the top of the ranking. 

Note also that using hits@10 to evaluate relational models is not ideal; a model can rank reasonable alternative entities  highly, but be penalized because the target entity is not in the top 10. For example, given ``rabbit IS-A'', both ``animal'' and ``mammal'' fit as target entities. This is alleviated by filtering, but is not completely eliminated due to the sparsity of relations in the dataset (which is the reason we require the link prediction task). Thus, we believe the mean rank is a more accurate measure of the performance of a model, particularly on raw data.

\begin{table}[t]
\footnotesize
\begin{center}
\begin{tabular}{ |l| }
  \hline
  \multicolumn{1}{|l|}{\textbf{WordNet Relations}} \\ \hline
  \_hyponym  \\ \hline
  \_derivationally\_related\_form \\ \hline 
  \_member\_holonym \\ \hline \hline
  \multicolumn{1}{|l|}{\textbf{Freebase Relations}} \\ \hline
  /award/award\_nominee/award\_nominations./award/ \\
    \hspace{1mm}
    award\_nomination/nominated\_for  \\ \hline
  /broadcast/radio\_station\_owner/ 
  radio\_stations  \\ \hline
  /medicine/disease/notable\_people\_with 
  \_this\_condition \\ \hline
\end{tabular}
\end{center}

\caption{ \label{relations} Sample relations from WordNet and Freebase. The relations from Freebase are clearly much more specific as they relate named entities.}
\end{table}

\paragraph{Dataset differences} It is also interesting to note the discrepancy between the results on the WordNet and Freebase datasets. Although using the entity descriptions leads to a significantly lower mean rank for the WordNet dataset, it only results in a faster convergence rate for Freebase. 
However, the relations presented in these two datasets are significantly different: WordNet relations are quite general and are meant to provide links between concepts, while the Freebase relations are very specific, and denote relationships between named entities. This is shown in Table~\ref{relations}. 
It seems that incorporating the definition of these named entities does not improve the ability of the algorithm to answer very specific relation questions. This would be the case if the optimization landscape for the TransE model had fewer local minima for Freebase than for WordNet, thus rendering it less sensitive to the initial condition. It is also possible that the TransE model is simply not powerful enough to achieve a filtered mean rank lower than 90, no matter the initialization strategy.


\section{Conclusion and Future Work}

We have shown that leveraging external lexical resources, along with distributional semantics, can lead to both a significantly improved optimum and a faster rate of convergence when applied with the TransE model for relational data. We established new state-of-the-art results on WordNet, and obtain small improvements to the state-of-the-art on raw relational data for Freebase. Our method is quite simple and could be applied in a straightforward manner to other models that take entity vector representations as input. Further research is needed to investigate whether performance on other NLP tasks can be improved by leveraging available lexical resources in a similar manner.

More complex methods initialization methods could easily be devised, e.g. by using inverse document frequency (idf) weighted averaging, or by applying the work of Le et al.~\shortcite{le2014distributed} on paragraph vectors. Alternatively, distributional semantics could be used as a regularizer, similar to \cite{labutov2013re}, with learned embeddings being penalized for how far they stray from the pre-trained GloVe embeddings. However, \textit{even with intuitive and straightforward methodology}, leveraging lexical resources can have a significant impact on the results of models for multi-relational data. 

\bibliography{acl2016}
\bibliographystyle{acl2016}

\end{document}